\documentclass{article}

\usepackage{PRIMEarxiv}
\usepackage{bm}
\usepackage{multirow}
\usepackage[utf8]{inputenc} 
\usepackage[T1]{fontenc}    
\usepackage{hyperref}       
\usepackage{url}            
\usepackage{booktabs}       
\usepackage{amsfonts}       
\usepackage{nicefrac}       
\usepackage{microtype}      
\usepackage{lipsum}
\usepackage{fancyhdr}       
\usepackage{graphicx}       
\graphicspath{{media/}} 
\usepackage{amsmath}
\title{Wormhole: Concept-Aware Deep Representation Learning for Co-Evolving Sequences

}

\author{
  Kunpeng Xu, Lifei Chen, Shengrui Wang \\
  Université de Sherbrooke \\
  Québec, Canada
}

\begin{document}
\maketitle

\begin{abstract}
Identifying and understanding dynamic concepts in co-evolving sequences is crucial for analyzing complex systems such as IoT applications, financial markets, and online activity logs. These concepts provide valuable insights into the underlying structures and behaviors of sequential data, enabling better decision-making and forecasting. This paper introduces \textit{Wormhole}, a novel deep representation learning framework that is concept-aware and designed for co-evolving time sequences. Our model presents a self-representation layer and a temporal smoothness constraint to ensure robust identification of dynamic concepts and their transitions. Additionally, concept transitions are detected by identifying abrupt changes in the latent space, signifying a shift to new behavior—akin to passing through a "wormhole". This novel mechanism accurately discerns concepts within co-evolving sequences and pinpoints the exact locations of these "wormholes,"  enhancing the interpretability of the learned representations. Experiments demonstrate that this method can effectively segment time series data into meaningful concepts, providing a valuable tool for analyzing complex temporal patterns and advancing the detection of concept drifts.
\end{abstract}


\section{Introduction}
With the rapid and continuous generation of data, understanding and analyzing dynamic concepts in sequential data is paramount for numerous applications, including IoT systems, financial market analysis, and online behavior monitoring \cite{matsubara2019dynamic,article,tao2023sqba,pang2024distributed,zhang2024learning,dan2024evaluation}. Such data often involve co-evolving sequences, where multiple time series exhibit interdependent behavior concepts over time. These co-evolving sequences encapsulate a wealth of information, revealing underlying structures and behaviors that are crucial for making informed decisions and predicting future trends.

Identifying concepts within these sequences offers significant advantages. For example, in IoT applications, concepts/patterns in sensor data can provide insights into operational efficiency and anomaly detection \cite{kawabata2019automatic,tao2023mlad,dan2024multiple}. In financial markets, understanding the co-movements of stock prices can aid in portfolio management and risk assessment \cite{huang2023generative}. Similarly, analyzing online activity logs can enhance user experience and targeted advertising \cite{ding2023temporal,10.1145/3331184.3331244,10653353,Ma_2024}. The challenge, however, lies in accurately identifying these concepts in real time, as data streams are often vast and continuously evolving. 

Concept identification in time series has been extensively studied, especially in database management and data mining \cite{fan2023dish,miyaguchi2019cogra,pang2023spectral,lam2024analyzing}. Traditional approaches such as hidden Markov models (HMM), autoregression (AR), and linear dynamical systems (LDS) are effective for static datasets but struggle with continuous data streams due to their lack of adaptability and need for predefined parameters \cite{dabrowski2018state,de2011forecasting}. Recent advancements in data stream mining, including CluStream \cite{aggarwal2003framework} and DenStream \cite{cao2006density}, offer improved scalability for evolving data but often fail to capture temporal dependencies and dynamic transitions. These methods focus more on maintaining clusters over time rather than understanding underlying concepts. Moreover, deep learning approaches like OneNet \cite{wen2024onenet} and FSNet \cite{pham2022learning} have made significant strides in forecasting by adapting to concept drift, yet primarily aim to enhance predictive accuracy without providing deeper insights into concept identification.

\begin{figure*}[!ht]
\centerline{\includegraphics[width=0.8\linewidth]{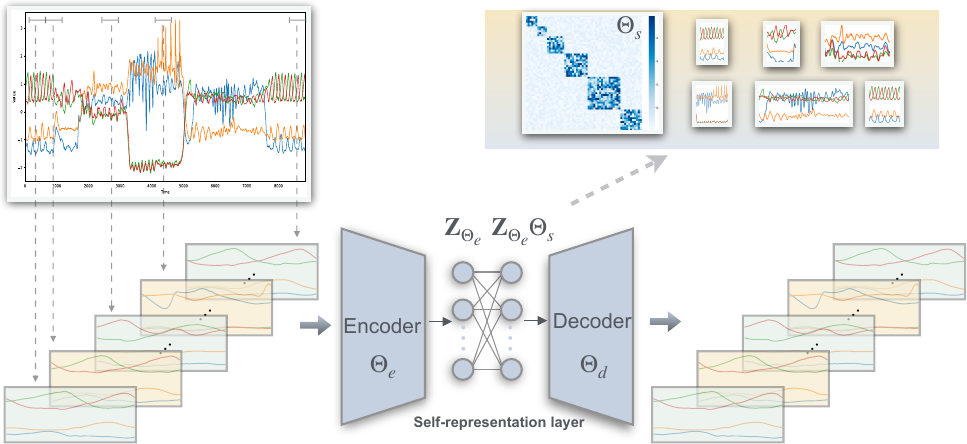}}
\caption{Framework}
\label{Fig1}
\vspace{-10pt}
\end{figure*}

To address these challenges, we propose a novel framework named \textit{Wormhole}, a concept-aware deep representation learning approach tailored for co-evolving time sequences. Our model leverages a self-representation layer, which effectively captures the intrinsic relationships among sequences, and a temporal smoothness constraint, which ensures that the transitions between identified concepts are coherent and meaningful. Unlike traditional methods, \textit{Wormhole} is designed to operate in a streaming context, even with large-scale co-evolving time series data, allowing for incremental detection and adaptation to new dynamic concepts as they emerge.

The key innovation of our approach lies in its ability to identify concept transitions by detecting abrupt changes in the latent space. These transitions, metaphorically described as “passing through a wormhole,” signify a shift to new behavior concepts and provide a clear demarcation of different concept segments. This mechanism not only enhances the interpretability of the learned representations but also allows for precise pinpointing of the transition points, offering valuable insights into the temporal evolution of the sequences.

We conducted experiments on various real-world datasets to evaluate the performance of our method. The results demonstrate that \textit{Wormhole} effectively segments time series data into meaningful concepts, outperforming traditional batch processing methods in terms of computational efficiency and the ability to handle co-evolving sequences. Furthermore, our approach shows significant improvements in detecting concept drifts, providing a powerful tool for analyzing complex temporal patterns in diverse applications.

\section{Related Work}

\subsection{Concept Drift in Time Series}

Concept drift has been a significant challenge in time series analysis, particularly in streaming data environments where the underlying data distributions may change over time~\cite{xu2024rhine,song23c_interspeech}. Traditional models such as Hidden Markov Models (HMM) and Autoregression (AR) have been widely used but often lack adaptability in the presence of continuous data streams. Recent advancements, such as OrbitMap~\cite{matsubara2019dynamic}, KRL~\cite{xu2024kernel} and TKAN~\cite{xu2024kolmogorov}, have improved scalability but still face challenges in capturing temporal dependencies and dynamic transitions. Additionally, models like Cogra~\cite{miyaguchi2019cogra} and Dish-TS~\cite{fan2023dish} have introduced techniques to address concept drift by incorporating stochastic gradient descent and distribution shift alleviation, respectively.

\subsection{Co-Evolving Sequences and Dynamic Concept Identification}

The identification of dynamic concepts in co-evolving sequences is crucial for understanding complex temporal patterns. Various methods have been proposed to segment time series data into meaningful patterns, including the use of hierarchical HMM-based models like AutoPlait~\cite{matsubara2014autoplait}, and the Toeplitz inverse covariance-based clustering method, TICC~\cite{hallac2017toeplitz}. Techniques have also been developed to analyze changes in mobility patterns caused by events such as COVID-19~\cite{zhang2021time}, Sequence pattern-based decision making~\cite{xu2023drnet}, financial market forecasting using clustering-based cross-sectional regime identification~\cite{chen2022clustering}, and dynamic cross-sectional regime identification for market prediction~\cite{chen2022dynamic}, highlighting the importance of adapting models to dynamic environments.

\subsection{Deep Representation Learning for Temporal Data}

Deep learning techniques have gained traction in temporal data analysis, offering powerful tools for representation learning. OneNet~\cite{wen2024onenet} and FSNet~\cite{pham2022learning} are notable examples that enhance time series forecasting by adapting to concept drift. However, these models primarily aim to improve predictive accuracy rather than offering insights into the concept identification process. Informer~\cite{zhou2021informer}, TIMESNET~\cite{wu2022timesnet}, Triformer~\cite{cirstea2022triformer}, and Non-stationary Transformers~\cite{liu2022non} further extend the capabilities of deep learning models in handling long sequence time-series forecasting and non-stationary behaviors in time series. Our work, Wormhole, builds on these ideas by introducing a self-representation layer that captures the intrinsic relationships among sequences and a temporal smoothness constraint that ensures coherent concept transitions. 

\subsection{Concept-Aware Models}

The idea of concept-aware models, which can detect transitions between different behaviors or patterns, has been explored in various domains~\cite{xu2022data,xu2022multi}. StreamScope~\cite{kawabata2019automatic} and the Generative Learning model~\cite{huang2023generative} for financial time series have contributed to this area by automatically discovering patterns in co-evolving data streams. There are also advancements in invariant time series forecasting in smart cities~\cite{zhang2024towards}, location-aware social network recommendations using temporal graph networks~\cite{zhang2023location}, and evolving standardization techniques for continual domain generalization~\cite{xie2024evolving}. Other approaches such as online boosting adaptive learning~\cite{yu2024online}, temporal domain generalization via concept drift simulation~\cite{chang2023coda}, and drift-aware dynamic neural networks~\cite{bai2022temporal} have also been proposed to handle concept drift in temporal data. However, these models do not explicitly address the interpretability of the learned representations, a gap that Wormhole seeks to fill by providing clear demarcations of concept transitions, enhancing the understanding of dynamic temporal patterns.

\subsection{Self-Representation Learning}

Self-representation learning has emerged as an effective approach for uncovering intrinsic relationships within data~\cite{xu2018self,chen2021self}. In self-representation models, each data point or instance is represented as a linear or nonlinear combination of other points within the dataset, allowing for a compact and interpretable representation of dependencies. This approach has been widely adopted in fields like subspace clustering and sparse coding due to its ability to uncover latent structures without relying on predefined labels.

By employing techniques such as sparse regularization and low-rank constraints, self-representation models can highlight the most significant dependencies between instances while ignoring irrelevant information.

\section{METHODOLOGY}

In this section, we introduce our novel framework, \textit{Wormhole}, designed for concept-aware deep representation learning in co-evolving time sequences. The model builds upon a self-representation deep learning approach and integrates a temporal smoothness constraint to effectively capture and understand dynamic concepts and their transitions. To handle the co-evolving sequences, we divide the original multivariate time series $\mathbf{S}$ into multiple segments using a sliding window. Each segment is treated as an input for the model. Therefore, $\mathbf{W} = { \mathbf{w}_1, \mathbf{w}_2, \ldots, \mathbf{w}_n }$ represents the collection of these segments, where each $\mathbf{w}_i$ is a multivariate time series segment containing information about various time steps and channels within the window. Our framework is illustrated in Fig.\ref{Fig1}.

\subsection{Deep Representation Learning}
Our framework is built upon a deep neural network architecture that facilitates learning robust representations of co-evolving sequences. The model includes the following key components:

\begin{enumerate}
\item \textbf{Encoder}: A deep neural network that maps the input time series segments into a latent space.
\item \textbf{Self-representation Layer}: This layer encodes the notion of self-representation, ensuring that each latent representation can be expressed as a combination of other latent representations, capturing the intrinsic relationships between the segments.
\item \textbf{Decoder}: This component reconstructs the time series segments from the latent representations, ensuring meaningful representation.
\end{enumerate}

\subsection{Self-representation in Time Series}
The self-representation layer plays a crucial role in capturing the relationships among different time series segments in the latent space. The concept of self-representation implies that each segment in the latent space can be expressed as a linear combination of other segments, effectively capturing the dependencies and similarities among the segments.

Mathematically, self-representation is defined as:

\begin{equation}
\mathbf{Z_{\Theta_e}} = \mathbf{Z_{\Theta_e}} \mathbf{\Theta_s}
\end{equation}

where $\mathbf{Z_{\Theta_e}}$ represents the latent representations of the time series segments, and $\mathbf{\Theta_s}$ is the self-representation coefficient matrix. Each column of $\mathbf{\Theta_s}$ corresponds to a segment that is expressed as a combination of other segments.

To enforce sparsity in the self-representation matrix $\mathbf{\Theta_s}$, we introduce $\ell_1$ norm regularization:

\begin{equation}
\mathcal{L}_{\text{self}}(\mathbf{\Theta_s}) = |\mathbf{\Theta_s}|_1
\end{equation}

This encourages the model to use a minimal number of latent components to represent each segment, effectively highlighting the most significant relationships and dependencies.

The self-representation property allows the model to learn a compact and interpretable representation of the co-evolving sequences, which is essential for identifying underlying dynamic concepts and their transitions.

\subsection{Temporal Smoothness Constraint}
The self-representation learning layer does not consider the information that is implicitly encoded into ordered data, such as the spatial or temporal relationships between data samples. For time series data, this sequential ordering should be reflected in the coefficients of the self-representation matrix $\mathbf{\Theta_s}$ so that neighboring segments are similar, i.e., $\bm{\theta_{s,i}} \approx \bm{\theta_{s,i+1}}$. To incorporate this property, we introduce a temporal smoothness constraint that penalizes large differences between consecutive columns of the self-representation matrix $\mathbf{\Theta_s}$, thereby ensuring that the latent representations vary smoothly over time.

We define the temporal smoothness constraint using a lower triangular matrix $\mathbf{R}$, which has $-1$ on the diagonal and $1$ on the second diagonal. This matrix effectively captures the differences between consecutive columns in $\Theta_s$.

\begin{equation}
\mathbf{R} =
\begin{bmatrix}
-1 & 1 & 0 & \cdots & 0 \\
0 & -1 & 1 & \cdots & 0 \\
\vdots & \vdots & \vdots & \ddots & \vdots \\
0 & 0 & 0 & \cdots & -1 \\
\end{bmatrix}
\end{equation}

Therefore, the product $\Theta_s \mathbf{R}$ can be represented as:

\begin{equation}
\mathbf{\Theta_s}{\mathbf{R}} = [\bm{\theta_{s,2}} - \bm{\theta_{s,1}}, \bm{\theta_{s,3}} - \bm{\theta_{s,2}}, \ldots, \bm{\theta_{s,n}} - \bm{\theta_{s,n-1}}]
\end{equation}

And the temporal smoothness constraint is defined as:

\begin{equation}
\mathcal{L}{\text{smooth}}(\Theta_s) = |\Theta_s \mathbf{R}|_{1,2}
\end{equation}

where $|\Theta_s \mathbf{R}|_{1,2}$ minimizes the column-wise $\ell{1,2}$ norm of $\Theta_{\mathbf{R}}$, encouraging smooth transitions between consecutive latent representations. This constraint ensures that changes in the dynamic concepts are captured effectively by penalizing large deviations in the latent space.

\subsection{Network Architecture}
The proposed model consists of an encoder, a self-representation layer, and a decoder. The encoder ($\mathbf{\Theta_e}$) maps the input time series segments $\mathbf{W}$ into a latent space representation $\mathbf{Z_{\Theta_e}}$:

\begin{equation}
\mathbf{Z_{\Theta_e}} = f_{\mathbf{\Theta_e}}(\mathbf{W})
\end{equation}

where $f_{\mathbf{\Theta_e}}$ represents a nonlinear function parameterized by $\mathbf{\Theta_e}$. The self-representation layer ($\mathbf{\Theta_s}$) enforces the self-representation property by ensuring that each latent representation can be expressed as a linear combination of other latent representations:

\begin{equation}
\mathbf{\hat{Z}_{\Theta_e}} = \mathbf{Z_{\Theta_e}} \mathbf{\Theta_s}
\end{equation}

The decoder ($\mathbf{\Theta_d}$) reconstructs the time series segments from the latent representations:

\begin{equation}
\hat{\mathbf{W}}_{\Theta} = g_{\mathbf{\Theta_d}}(\mathbf{\hat{Z}_{\Theta_e}})
\end{equation}
where $g_{\mathbf{\Theta_d}}$ denotes the decoding nonlinear function parameterized by $\mathbf{\Theta_d}$. It transforms the latent space representations back into the original input space, reconstructing the time series segments. The complete loss function for the model combines reconstruction loss, self-representation regularization, and temporal smoothness penalty:

\begin{equation}
\mathcal{L}(\mathbf{\Theta}) = \frac{1}{2} |\mathbf{W} - \hat{\mathbf{W}}_{\Theta}|_F^2 + \lambda_1 |\mathbf{\Theta_s}|1 + \lambda_2 |\mathbf{Z_{\Theta_e}} - \mathbf{Z_{\Theta_e}} \mathbf{\Theta_s}|F^2 + \lambda_3 |\mathbf{\Theta_s} \mathbf{R}|_{1,2}
\end{equation}

Here, $\mathbf{\Theta}$ encompasses the parameters of the encoder ($\mathbf{\Theta_e}$), the self-representation layer ($\mathbf{\Theta_s}$), and the decoder ($\mathbf{\Theta_d}$). The parameters $\lambda_1$, $\lambda_2$, and $\lambda_3$ are regularization coefficients that control the balance between the reconstruction accuracy, sparsity of the self-representation matrix, and temporal smoothness. Specifically:
- $\lambda_1$ controls the sparsity of the self-representation matrix $\Theta_s$, promoting a sparse representation.
- $\lambda_2$ ensures that the self-representation property holds by minimizing the difference between $\mathbf{Z_{\Theta_e}}$ and $\mathbf{Z_{\Theta_e}}  \mathbf{\Theta_s}$.
- $\lambda_3$ enforces temporal smoothness by minimizing the deviations in the temporal difference matrix $ \mathbf{\Theta_s} \mathbf{R}$.

\subsection{Concept Transition Detection}
The process of concept transition can be likened to jumping through a wormhole in space, where the boundaries between different concepts resemble wormholes. These boundaries mark the points of transition from one concept to another, facilitating a rapid change in behavior patterns.

To determine the locations of these wormholes, or boundaries, we analyze the distribution of the self-representation matrix $ \mathbf{\Theta_s} \mathbf{R}$. Significant changes in $ \mathbf{\Theta_s} \mathbf{R}$ indicate shifts in the underlying behavior patterns. This method uses the temporal differences captured in $ \mathbf{\Theta_s} \mathbf{R}$ to identify transitions, where large deviations suggest a boundary between different concepts or segments in the data.

\begin{figure}[!t]
\centerline{\includegraphics[width=0.6\linewidth]{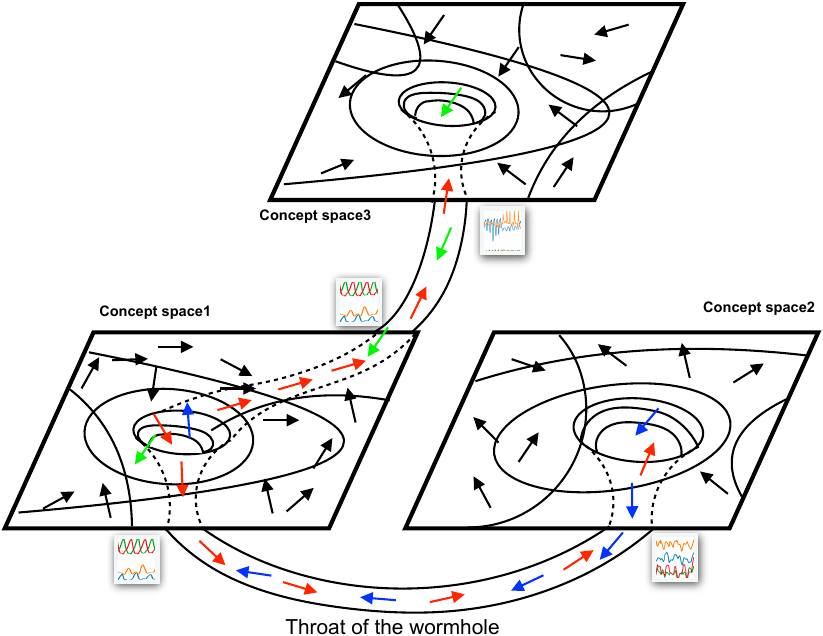}}
\caption{Concept transition through "wormhole".}
\label{Fig2}

\end{figure}

If we assume that the series is drawn from a set of disconnected concept spaces (i.e., $ \mathbf{\Theta_s}$ is block diagonal), the information encoded by $ \mathbf{\Theta_s} \mathbf{R}$ can reveal the space boundaries. Ideally, the columns of $ \mathbf{\Theta_s} \mathbf{R}$, i.e., $\bm{\theta}_{s,i} - \bm{\theta}_{s,i-1}$, that are within a segment should be close to the zero vector because columns from the same concept subspace share similarity. Columns of $ \mathbf{\Theta_s} \mathbf{R}$ that greatly deviate from the zero vector indicate the boundary of a segment, akin to transitioning through a wormhole to a new concept space, as illustrated in Fig.\ref{Fig2}.

First, we compute the absolute value matrix of $\mathbf{\Theta_s} \mathbf{R}$, i.e., $\mathbf{B} = |\mathbf{\Theta_s} \mathbf{R}|$. Then, we calculate the column-wise means of $\mathbf{B}$ -- $\mathbf{y} = \text{mean}(\mathbf{B}, \text{axis}=1)$. We employ a peak-finding algorithm over $\mathbf{y}$ to identify the segment boundaries. Peaks in $\mathbf{y}$ correspond to the points where the segments likely transition to a new concept.

This intrinsic segmentation method effectively detects changes in the underlying structure of the time series, facilitating the identification of dynamic concept transitions by pinpointing the exact locations of these wormholes.

\section{Experiments}
In this section, we present the experiments conducted to evaluate the effectiveness of the proposed \textit{Wormhole} framework for concept-aware deep representation learning in co-evolving time sequences. We describe the datasets used, the experimental setup, the evaluation metrics, and the results.

\subsection{Datasets}
We evaluated our model using three datasets representing different domains of co-evolving time series. The Motion Capture Streaming Data from the CMU database\footnote{\url{MoCap: http://mocap.cs.cmu.edu/}} captures various motions such as walking and dragging, making it ideal for analyzing transitions between different types of human activities. The Stock Market Data includes historical prices and financial indicators from 503 companies\footnote{\url{https://ca.finance.yahoo.com/}}, providing a large-scale, high-dimensional dataset to test the model's performance in detecting concept changes in financial markets. Lastly, the Online Activity Logs from GoogleTrend event streams\footnote{\url{http://www.google.com/trends/}} include 20 time series of Google queries for a music player from 2004 to 2022, used to evaluate the model's ability to detect behavioral shifts in user interactions.

\subsection{Comparison with Baseline Models}
To thoroughly evaluate the effectiveness of our \textit{Wormhole} model, we compared it with several baseline models. These included StreamScope\cite{kawabata2019automatic}, a scalable streaming algorithm for automatic pattern discovery in co-evolving data streams; TICC\cite{hallac2017toeplitz}, which segments time series into interpretable clusters based on temporal dynamics; and AutoPlait\cite{matsubara2014autoplait}, a hierarchical HMM-based model for automatic time series segmentation that identifies high-level patterns. The experimental results are summarized in Table \ref{tab}.

\begin{table}[!t]
\caption{Comparison of \textit{Wormhole} and baseline models.}
\label{tab}
\resizebox{1\textwidth}{!}
{
\begin{tabular}{|c|c|c|c|c|c|}
\hline
\textbf{Dataset} & \textbf{Metric} & \textbf{StreamScope} & \textbf{TICC} & \textbf{AutoPlait} & \textbf{Wormhole} \\ \hline

\multirow{2}{*}{\textbf{Motion Capture Data}} & \textbf{F1-Score} & 0.81 & 0.43 & 0.84 & \textbf{0.86} \\ \cline{2-6}
 & \textbf{ARI} & 0.55 & 0.19 & 0.54 & \textbf{0.59} \\ \hline

\multirow{2}{*}{\textbf{Stock Market Data}} & \textbf{F1-Score} & 0.70 & 0.28 & 0.72 & \textbf{0.82} \\ \cline{2-6}
 & \textbf{ARI} & 0.58 & 0.18 & 0.71 & \textbf{0.77} \\ \hline

\multirow{2}{*}{\textbf{Online Activity Logs}} & \textbf{F1-Score} & 0.87 & 0.76 & 0.86 & \textbf{0.88} \\ \cline{2-6}
 & \textbf{ARI} & 0.83 & 0.72 & 0.81 & \textbf{0.86} \\ \hline

\end{tabular}}

\end{table}

Table \ref{tab} demonstrates that our model consistently outperforms the baseline models across all datasets. Notably, \textit{Wormhole} achieves the highest F1-Score and ARI in both Motion Capture Data and Online Activity Logs, indicating its superior capability in detecting and segmenting concept transitions. In the Stock Market Data, despite the complexity and high dimensionality, \textit{Wormhole} significantly outperforms the baselines, highlighting its robustness in handling large-scale co-evolving time series.

\subsection{Visualization of Concept Transitions}
\begin{figure}[!t]
\centerline{\includegraphics[width=0.9\linewidth]{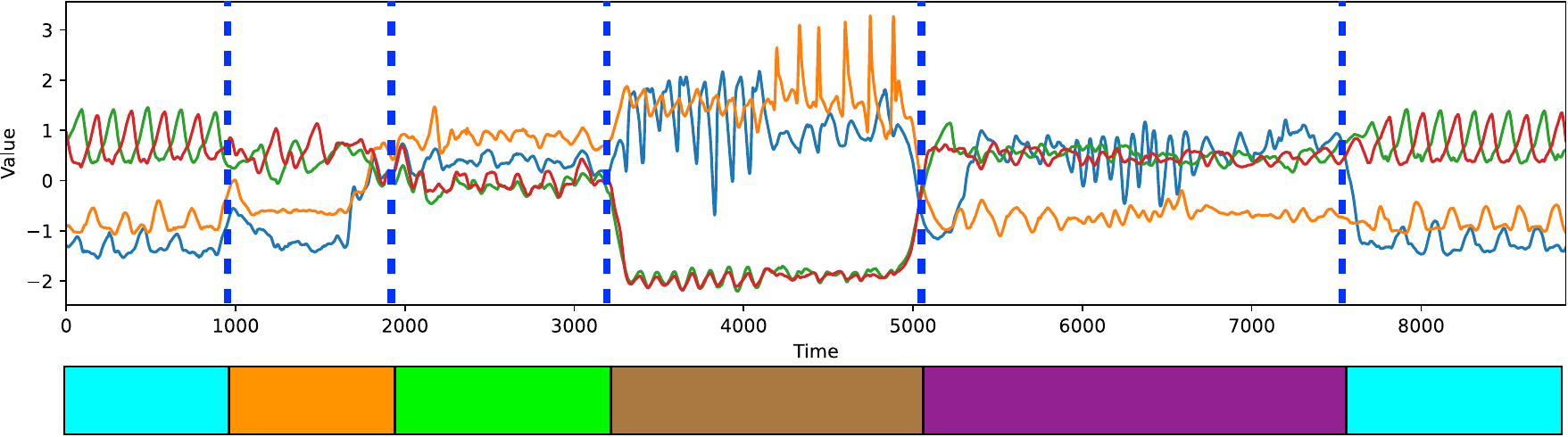}}
\caption{Identify concepts and transitions on motion data.}
\label{Fig3}

\end{figure}

We visualized the detected concept transitions for the Motion Capture Data. Fig. \ref{Fig3} shows the time series segments with marked concept transitions, highlighting how the model accurately identifies the boundaries between different types of motion such as walking and dragging. This visualization reinforces the effectiveness of the \textit{Wormhole} framework in detecting dynamic changes in co-evolving sequences.

\section{Conclusion}
We introduced \textit{Wormhole}, a framework for concept-aware deep representation learning in co-evolving time sequences. \textit{Wormhole} offers a promising approach for understanding complex temporal patterns, with potential applications in various domains. Our experiments show that it effectively detects and segments dynamic transitions across diverse datasets, outperforming current state-of-the-art methods. As future work, we aim to develop \textit{Wormhole} into a standalone module that can be integrated with advanced time series forecasting models to mitigate the impact of concept drift. This will enhance the robustness and accuracy of predictions in dynamically changing environments.

\bibliographystyle{elsarticle-num}  
\bibliography{references}

\end{document}